\def\year{2021}\relax
\documentclass[letterpaper]{article} 
\usepackage{aaai21}  
\usepackage{times}  
\usepackage{helvet} 
\usepackage{courier}  
\usepackage[hyphens]{url}  
\usepackage{graphicx} 
\def\UrlFont{\rm}  
\usepackage{comment}
\usepackage{natbib}  
\usepackage{caption} 
\usepackage{graphicx}
\title{Tile Embedding: A General Representation for Level Generation}
\author{
    Mrunal Jadhav and Matthew Guzdial\\
}

\affiliations{
    Department of Computing Science, Alberta Machine Intelligence Institute (Amii)\\
    University of Alberta, Canada\\
    \{mrunalsu, guzdial\}@ualberta.ca

}

\begin{document}
\maketitle
\begin{abstract}
In recent years, Procedural Level Generation via Machine Learning (PLGML) techniques have been applied to generate game levels with machine learning.
These approaches rely on human-annotated representations of game levels. Creating annotated datasets for games requires domain knowledge and is time-consuming. Hence, though a large number of video games exist, annotated datasets are curated only for a small handful.  
Thus current PLGML techniques have been explored in limited domains, with Super Mario Bros. as the most common example. 
To address this problem, we present tile embeddings, a unified, affordance-rich representation for tile-based 2D games. 
To learn this embedding, we employ autoencoders trained on the visual and semantic information of tiles from a set of existing, human-annotated games.
We evaluate this representation on its ability to predict affordances for unseen tiles, and to serve as a PLGML representation for annotated and unannotated games.
\end{abstract}
\section{Introduction}

Procedural Content Generation via Machine Learning (PCGML) is the generation of game content using Machine Learning (ML) models trained on existing data \cite{summerville2018procedural}. 
Unlike when machine learning is applied to image generation, PCGML models cannot train only on raw pixel representations of game content like levels.  
Game content obeys structural and functional constraints to ensure playability. 
This creates a need for secondary representations that capture the behaviour of game objects outside of pixels alone. 
Converting level screenshots to a parseable format requires annotations by experts or fan communities \cite{summerville2016vglc}. 
Thus PCGML applications are limited to a handful of domains where annotations exist. 

A level is a space the player travels through, interacting with objects like enemies and collectibles. 
A common approach to representing a game level is to map the pixel representation of game objects to a set of characters called tiles.
Each tile is associated with a set of in-game affordances.
Affordances convey the conceptual idea of the object and capture the possible interactions of the player with the object \cite{bentley2019videogame}. 
For instance, the Goomba \includegraphics[scale=0.6]{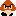} in Super Mario Bros(SMB) has the affordances \textit{Enemy, Damaging, Hazard, Moving} in a common PCGML representation \cite{summerville2016vglc}. 
Curating these datasets requires manual effort and the representations are game-specific.
Consider the problem of training a PCGML model for generating levels of the game Bubble Bobble. 
Since no annotated representation of its levels exists, we would have to parse the levels ourselves. 
This typically involves a series of tasks including processing images with OpenCV, human editing, extracting a reduced set of representative tiles, and tagging them with appropriate affordances based on their behaviour \cite{summerville2016vglc}. 
This represents a significant amount of work.

While each tile character is mapped to a set of affordances, the affordances are not directly included in the representation. 
For instance, the Goomba \includegraphics[scale=0.6]{./E.png} is represented with character \textit{‘E’} in the above-mentioned representation, and affordance-mapping is present in a separate JSON file.  
Hence, at their core, the level generation tasks that leverage these representations address problems as a character generation process. 
Appropriate visual reconstruction also impacts the choice of tiles to include. This enforces the requirement of position-specific tags in the affordance set of the tile/character. 
For instance, in SMB there are repeated pipe objects of different heights. They are often represented with four different tiles, ‘[’, ‘]’, ‘$<$’, ‘$>$’ representing the bottom-left \includegraphics[scale=0.6]{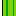}, bottom-right \includegraphics[scale=0.6]{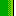}, top-left \includegraphics[scale=0.6]{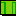} and top-right \includegraphics[scale=0.6]{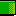} of a pipe respectively. 
In other instances, PCGML practitioners must author secondary processes to visualize levels, such as mapping different characters/tiles to different images depending on their y-position \cite{summerville2016vglc}.

The current, state-of-the-art PCGML level representation has a number of drawbacks, requiring substantial human effort when collecting data, game-specific representations, and extra processing to visualize generated levels.
To address these challenges, we present a domain-independent, affordance-rich representation of game levels, reducing the reliance on manual translations and domain expertise. We draw inspiration from word embeddings \cite{mikolov2013efficient} and present \textit{tile embeddings}, which integrate visual and semantic information.\footnotemark
\\
\\
The main contributions of this paper are :
\begin{enumerate}
  \item We propose tile embeddings as a representation for Procedural Level Generation via Machine Learning (PLGML).
  \item We demonstrate that our tile embeddings can approximate the affordances of tiles from unseen games.
  \item We apply this representation to level generation and demonstrate equivalent or better performance compared to the current, state-of-the-art representation.
  \item We demonstrate the ability to apply our representation to level generation for an unseen game, allowing for PLGML applications for any tile-based 2D game. 
\end{enumerate}

\footnotetext{The implementation is available at https://github.com/js-mrunal/tile\_embeddings}

\section{Related Work}
In this section we discuss autoencoders and current level representation practices in PCGML, overview word embeddings, and cover prior work on game embeddings. 

\subsubsection{Autoencoders} Prior research has successfully employed autoencoders \cite{hinton2006reducing} and Variational Autoencoders (VAEs) \cite{kingma2013auto} for PCGML tasks \cite{thakkar2019autoencoder, sarkar2020controllable}. An Autoencoder is an unsupervised model that learns to transform data into a compact vector representation while a Variational Autoencoder (VAE) maps data to a probabilistic distribution.
Jain et al. \cite{jain2016autoencoders} were the first to demonstrate that autoencoders could learn representations useful in downstream PCGML tasks. 
Guzdial et al. \cite{guzdial2018explainable} presented an explainable co-creative tool using an autoencoder by learning existing level structures and associated design pattern labels. 
While these works do not directly focus on embeddings, the essence of our approach is in learning and optimizing a latent representation. 
Alvernaz and Togelius \cite{alvernaz2017autoencoder} trained an autoencoder to generate a low dimensional representation of a videogame environment which was then used in a reinforcement learning framework.
This is similar to our approach as we learn a level representation using level structure and affordances. However, it focused on automating gameplay while our focus is automating design. Additionally, all these previous approaches are based on representations of chunks of levels or entire levels. In our work, we focus on a representation for tiles, a level's building blocks.

\subsubsection{Level Representation}Most PCGML approaches addressing level design tasks rely on datasets of annotated images  \cite{summerville2016super,snodgrass2017procedural,beaupre2018design,sarkar2020controllable}, or gameplay videos \cite{guzdial2016game}. A notable contribution to the current Game AI research community is the Video Game Level Corpus (VGLC) \cite{summerville2016vglc}. It presented a training corpus for 12 games consisting of level images and parseable text files in three different formats: tiles, graphs, and vectors. This work has gained popularity with 77 citations at the time of this writing. 

How one determines a set of affordances for tilesets is an open area of research. While most PCGML approaches rely on the hand-authored set from the VGLC or similar representations, there has been some effort to try to derive these in a more grounded way. Summerville et al. \cite{summerville2017does} attempted to learn the semantic properties of tiles from gameplay.
Snodgrass \cite{snodgrass2018towards} clustered potential tiles into groups and estimating their quality based on levels generated using these potential tiles. For clustering, the neighbouring tiles surrounding a tile played an important role. We incorporate a similar concept by training on the combination of the visual context of neighbouring tiles with the affordances from the VGLC. For level blending tasks \cite{sarkar2020exploring}, which combine different game representations, there’s a need to come up with a joint set of affordances across games, but this is typically done by hand. Bentley and Osborn \cite{bentley2019videogame} presented an annotation tool and a common set of nine affordances. We leverage the affordances from this tool.

\subsubsection{Embedding Vectors} Word embeddings \cite{mikolov2013efficient} are extensively used in modern NLP tasks. 
Each word is represented as a continuous d-dimensional vector denoted by $ w^i \in\ R^d$. 
The low-dimensional representation captures the word's meaning (semantics) from streams of text. Words related to each other are placed closer in the vector space, and relationships between words are encoded as the differences between these points. 
A popular word analogy that can be demonstrated by this vector space is $ \vec{king}\ -\ \vec{man}\ +\ \vec{woman}\ \approx\ \vec{queen}$, which demonstrates an understanding of concepts and context by the model. 

World models represent a novel approach to learning to represent an entire game or similar virtual environment as a neural network \cite{ha2018world,kim2020learning}.
Related to this,
Yousefzadeh Khameneh and Guzdial \cite{khameneh2020entity} used a VAE to extract embeddings of the entities in a game, which they call \textit{entity embeddings}, which encoded information of gameplay elements. We instead focus on capturing the level structure in our representation and define \textit{tile embeddings} as a d-dimensional vector in an embedding space encoding the semantic information of a tile.
\begin{figure*}[tbh]
    \centering
     \includegraphics[width=\textwidth,scale=0.7]{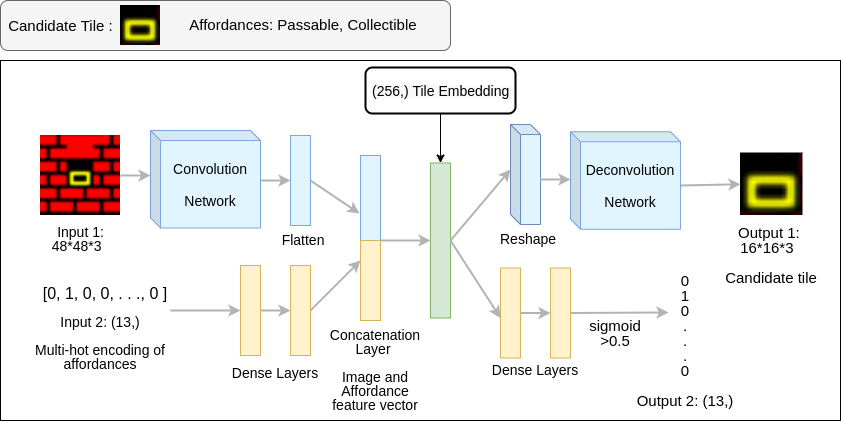} 
     \caption{Network Architecture.}
     \label{fig:Example2}
\end{figure*}

\section{System Overview}
The goal of our work is to learn an affordance-rich embedding of a tile as a Procedural Level Generation via Machine Learning (PLGML) representation. 
We employ the widely-used autoencoder framework to learn this embedding. Our training data consists of five classic Nintendo games- Super Mario Bros, Kid Icarus, Legend of Zelda, Lode Runner and Megaman, which are all 2D, tile-based games. Figure \ref{fig:Example2} illustrates our architecture, described in detail below.

\begin{figure}[tb]
    \centering
     \includegraphics[scale=0.6]{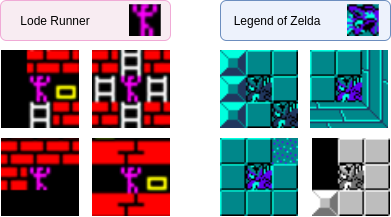} 
     \caption{Neighbourhood Context for Tiles.}
     \label{fig:neighbourhood}
\end{figure}

We draw on local pixel context and affordances associated with the tiles from the VGLC \cite{summerville2016vglc}. We incorporate affordances as an input since the visual similarity between tiles can be deceptive. Tiles that differ in pixel appearance may have the same behaviour, such as the recoloured tiles in Figure \ref{fig:neighbourhood}. 
Further, when affordances are not known, the neighbourhood context could be crucial for the embedding vector. For instance, a brick may depict a [‘solid’, ‘breakable’] object in one game, but a background pattern in another game with the affordances [‘empty’, ‘passable’]. However, in this case, the bricks in the latter case would repeat in a way similar to solid coloured sky tiles in other games. Thus the placement of a tile's embedding value in the latent space is influenced not only by the visuals of the tile but also by its behaviour and relationship with the neighbouring tiles. 

\subsection{Local Pixel Context}
One part of the input to our autoencoder is the pixel representation of a tile and its neighbouring tiles as demonstrated in Figure \ref{fig:neighbourhood}. 
For this, we use the level images from the VGLC Corpus. 
To capture local context, we slide a 48*48 pixel window, as we use a 16*16 tile representation, over the images to extract all unique contexts.
By unique we indicate all possible combinations of VGLC tile types in the neighbourhood of the candidate tile, it does not matter if the tiles differ in terms of their pixel appearance. We made this choice to reduce class imbalance in tile types, as ``empty'' background tiles occur much more frequently than all others.

\subsection{Semantic Context and Unified Affordances}
The other input to our model is the affordances of the candidate tile. The annotations for each of the tiles are obtained from the JSON files stored in the VGLC Corpus \cite{summerville2016vglc}. However, these are all game-specific, thus it is necessary to map the different game affordances to a single, unified set. 
Based on prior work \cite{bentley2019videogame,sarkar2020exploring}, we employ the following 13 common tags: \textit{Block, Breakable, Climbable, Collectable, Element, Empty, Hazard, Moving, Openable, Passable, Pipe, Solid, Wall}. For example, \textit{Climbable, Passable} refers to tiles such as stairs, ropes, and ladders. The player can use these tiles to move in the vertical direction or can choose to pass the tile and continue on their original path. \textit{Hazard} covers all harmful obstacles to the player such as spikes, cannons, and enemies. The affordances for each tile are then expressed as a multi-hot vector, with 1 at the index of features that are present for this tile, and 0 otherwise. 

\subsection{Model Architecture}
An autoencoder is a feedforward multilayer neural network architecture that learns a compressed representation of the input to capture key structures. In our work, we adapt the X-Shaped VAE architecture proposed by Simidjievski et al.  \cite{simidjievski2019variational}. The encoder consists of two branches that process the individual inputs. The outputs of the two branches are merged and compressed into a single embedding vector which we employ as our tile embedding. The decoder network again splits into two branches to reconstruct the desired outputs.  

The 48*48 pixel input is fed to a three-layer encoder convolutional network - the first with 32 (3*3), then 32 (3*3) and finally 16 (3*3) filters. Each layer is followed by Batch Normalization and then Tanh Activation. Batch normalization applied before a non-linear activation function stabilizes the distribution of the input and reduces the divergence risk \cite{ioffe2015batch}. 
This output is flattened to form a one-dimensional image feature vector. In parallel, the multi-hot feature vector of affordances is passed through two fully connected layers of sizes 32 and 16 with Tanh activation for a feature vector encoding of the affordances. 

We concatenate the output of both branches and pass it through a fully connected layer to get a (256,) dimensional tile embedding. This captures the relationships between branches in a common latent representation. This merging of information is crucial in cases where the affordance information is unknown, such as when we wish to derive tile embeddings for a new game. We hypothesize in these cases that we can approximate reasonable affordances based on pixel data alone. 

The decoder is close to an inverse of the encoder. A three-layer deconvolutional network upsamples the embedding vector to reconstruct the pixel portion of our output. 
Given that we want an embedding for individual tiles, we reconstruct just the 16*16 centre tile. 
In parallel, in order to reconstruct the affordances, we include two fully connected layers of sizes 16, and 32. The output of these layers is finally connected to a dense layer with Sigmoid activation representing the affordances of the centre tile. 

We trained this model with the adam optimizer and two-loss functions. 
For the image output, we use mean square loss. 
The multi-label prediction task for our $N$ affordances can be formulated as $N$ independent binary classification problems and so we use binary cross-entropy loss as our second loss function. 
However, our training dataset does not have equal instances of each label. To counter this problem of class imbalance, we derived a TF-IDF vectorizer to compute an importance score for each label based on its frequency. We use this as the weight for each label and define our binary cross-entropy as, 
\begin{equation}
Weighted\ BCE = -\sum_{i=1}^{N} y_i log(P(y_i)) * w_i
\end{equation}
where, $y_i$ is the ground truth, $P(y_i)$ is the predicted probability for label $i$ in $N$, and $w_i$ is the TF-IDF weight for label $i$. The objective function combines the two above loss functions with a weighted linear combination. We use the weight 0.8 for the image loss and 0.2 for the affordance loss, which we derived empirically. During training, we employ 20\% of our training data as a validation set and apply early stopping to avoid overfitting \cite{prechelt1998early, hawkins2004problem}. 

\begin{figure}[tb]
    \centering
     \includegraphics[width=\columnwidth,scale=0.45]{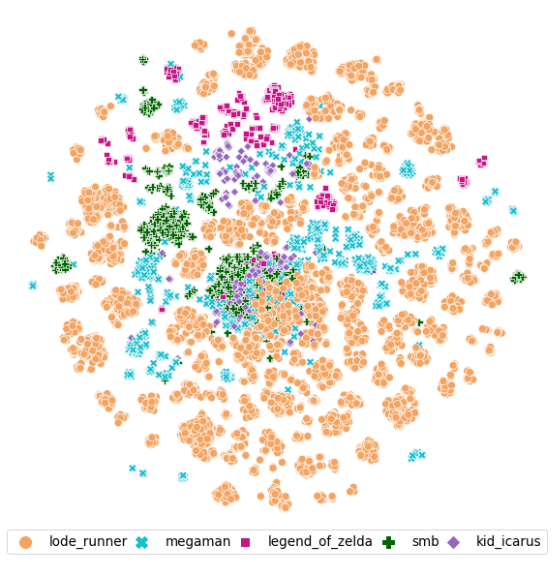} 
     \caption{t-SNE Visualization of Embedding Space.}
     \label{fig:tsne}
\end{figure}

We include a t-SNE \cite{van2008visualizing} visualization of our learned latent space (Figure \ref{fig:tsne}). It shows a good mix of our tile embeddings across different games. 
Lode Runner is over-represented as it has the most samples of any game. 
However, even games like Legend of Zelda, which are very different from the other games, are fairly evenly distributed across this latent space, indicating it has been able to generalize across the different games. 

\section{Evaluation}
In this section, we discuss the three evaluations of our system. 
First, we approximate affordances for tiles of unseen games. 
Second, we compare tile embeddings and the VGLC tile representation on a level generation task.
Finally, we demonstrate the application of tile embeddings for generating levels of a game with no annotated data.

\subsection{Cross-fold Affordances Analysis}
We employ a cross-fold analysis over our five games: Super Mario Bros, Kid Icarus, Legend of Zelda, Lode Runner, and Megaman. Our model is trained on four games with the fifth game held out as test data. We extract and pass 48*48 pixel contexts from test levels as the input to our trained model. 
We act as though their affordances are not known and pass a (13,) array of zeros as the second input.
This allows us to approximate a situation in which we are attempting to predict the affordances for an unseen game.

Evaluating the predicted affordances is a multi-label prediction task where the predicted output may be fully correct, partially correct, or fully incorrect. 
We therefore employ a number of metrics.
\textit{Exact Matching Ratio (EMR)} indicates the percentage of test examples where the predicted labels are exactly correct. EMR can be harsh in a multi-label setting. Hence we adopt example-based and label-based evaluations from \cite{sorower2010literature} with the metrics: \textit{Precision, Recall, Accuracy} to evaluate our model for partial correctness. 
We include \textit{Example-based} versions of these metrics, which are applied on each instance and averaged over the number of instances in the dataset.  
For the \textit{Label-based} version of these metrics, we investigate their values for individual labels and compute the average on each label’s precision, recall and accuracy independently. 
The example-based metrics allow us to determine our performances in terms of all the labels (affordances) of each tile, whereas the label-based metrics capture the performance in terms of individual labels (empty, hazard, etc.).
Accuracy provides an intuition of the model’s correctness in predicting true positives(TP) and true negatives(TN). However, for a sparse prediction vector, accuracy may be misleading. To understand the performance of the model at predicting positives accurately, we employ Precision and Recall. Of all the labels that the model predicted (TP+FP), precision indicates the percentage of labels that were actually true (TP). On the other hand, Recall is the percentage of true labels that the model was able to capture (TP/ (TP+FN)). To further investigate misclassification and missing-label errors, we adopt a more robust metric: \textit{Alpha Evaluation} \cite{boutell2004learning}. Alpha Evaluation weights missing-label errors ($M_x$) and misclassification errors ($F_x$) separately using parameters $\beta$ (for missing-label) and $\gamma$ (for misclassification). $\alpha$ controls the forgiveness for errors. Alpha Evaluation is given by the formula,
\begin{equation}
    alpha\ score=(1-\frac{|\beta M_x\ +\ \gamma F_x|}{|Y_x\ \vee\ P_x|})^\alpha
\end{equation}
such that $ \alpha \geq 0,\ 0\leq\ \beta, \beta=1|\gamma=1,$ where $\ Y_x$ is the ground truth and $P_x$ are the predicted labels. 

\subsection{LSTM for Level Generation-Annotated Game}

In this evaluation, we directly compare our tile embedding representation to the state-of-the-art VGLC representation for one game. 
LSTMs are a special type of RNN with a memory mechanism at the heart of their architecture. 
LSTMs have been extensively used in PLGML. We adapt the work of Summerville and Mateas \cite{summerville2016super} and train two similar LSTM networks, one with the VGLC tile representation and the other with our tile embeddings to generate levels for the game Lode Runner. 
We chose Lode Runner due to the results of the first evaluation.
Lode Runner tiles are 8*8 pixels in size. To fit this to our autoencoder architecture, we upscaled the level images using the Python Imaging Library (PIL) such that each tile has a dimension of 16*16 pixels.
We trained our model to consider a history of the last 3 rows (approximately 100 tiles) and generate the next 3 rows at a time. Similar to Summerville and Mateas' work, to track the progression of the level, we include column depth as an input to the network. The only differences between the two network implementations are in the input and output layer due to the differences in representation.

\textbf{Input Layer}: Before training an LSTM on tile embeddings, each level is converted to an embedding representation with our trained autoencoder model using context windows and affordances. For instance, a (512 * 352*3) level image of Lode Runner is converted to a (32*22*256) representation.  The other LSTM is trained on the (32*22) character representation obtained from the VGLC dataset.

\textbf{Output Layer}: For the LSTM trained on our embedding representation, the output layer predicts the embedding directly. It is modelled as a (256,) Dense layer with Tanh activation. Before visualizing a level, we map the predicted embedding to the nearest actual embedding. 
We use the memory efficient Annoy library\footnote{https://github.com/spotify/annoy} to index the embeddings and find the nearest neighbor based on the Manhattan distance. For the VGLC representation, the output of the LSTM is connected to a dense layer with Softmax activation indicating the probability of a tile character. We perform an expressive range analysis of generated levels with the metrics: Linearity and Leniency \cite{summerville2018expanding, smith2010launchpad, marino2015empirical}.
\begin{itemize}
  \item \textit{Linearity} profiles the structure of a level in terms of how well it fits to a line. Linearity is computed by performing linear regression on centre points of all the platforms. We then compute the average distance between each centre point and its projection on the regression line. The score is normalized between [0,1] by dividing by the total number of centre points. 
  \item \textit{Leniency} measures the difficulty of the level. We assign rewards with weight 1 and enemies with weight -1. We then calculate the sum of leniency values and average it with the total number of tiles. 
\end{itemize}
\subsection{LSTM for Level Generation- Unannotated Game}
As our third evaluation, we apply tile embeddings for generating levels for the game Bubble Bobble. We chose this game because no annotated dataset for it currently exists. We download 100 Bubble Bobble level images as our training dataset.\footnote{https://www.adamdawes.com/retrogaming/bbguide/} We extract tile embeddings by passing the 48*48 pixel context and a (13,) zero vector to an autoencoder trained on all five NES games. 
We employ the same architecture as we did for Lode Runner to train an LSTM on the embedded level representation. The majority of the Bubble Bobble levels are vertically symmetric, and so we parse the levels column-wise. Our model is trained to generate the right half of the level when the left half is fed as an input. 
During inference, we mirror the generated right half to produce an entire level.

\section{Results}
\renewcommand{\arraystretch}{1.2}
\begin{table*}[!tbh]
\centering
\small
\begin{tabular} {|l|c c c c|c c c|c c c|}
\hline
Test Data & 
\multicolumn{4}{c|}{{Example-based}} &
\multicolumn{3}{c|}{Label-based} &
\multicolumn{3}{c|}{$\alpha$-Evaluation with $\alpha$=1} \\ 
& EMR     & Prec    & Recall    & Acc    & Prec    & Recall    & Acc   & $\beta$=0.75, $\gamma$=0.25 & $\beta$=1,$\gamma$=1 & $\beta$=0.25,$\gamma$=0.75\\ 
\hline
SMB & 0.17 & 0.52 & 0.49 & 0.39 & 0.22 & 0.23 & 0.11 & \textit{0.66} & 0.29 & 0.63 \\
Kid lcarus  & \textbf{0.44} & 0.63 & 0.55 & 0.54 & 0.27 & 0.30 & \textbf{0.14} & 0.75 & \textbf{0.45} & 0.69  \\ 
Megaman     & 0.36 & 0.60 & \textbf{0.61} & 0.53 & \textit{0.25} & \textbf{0.32} & \textbf{0.14} & 0.71 & 0.40 & \textbf{0.70} \\ 
Lode Runner & \textit{0.11} & \textit{0.44} & \textit{0.27} & \textit{0.27} & 0.26 & \textit{0.17} & \textit{0.05} & 0.67 & \textit{0.17} & \textit{0.50}  \\ 
LOZ         & 0.39 & \textbf{0.78} & \textbf{0.61} & \textbf{0.59} & \textbf{0.34} & \textit{0.17} & 0.10 & \textbf{0.78} & 0.43 & 0.65 \\ \hline
\hline
Mean & 0.29 & 0.59 & 0.51 & 0.46 &0.27 & 0.23 &0.11 & 0.71 & 0.35 & 0.63 \\ \hline
MFL Baseline & 0.32 & 0.46 & 0.46 & 0.42 & 0.46 & 0.15 & 0.07 & 0.64 & 0.30 & 0. 67 \\ \hline
\end{tabular}
\caption{Results of evaluation metrics for predicting affordances on unseen tiles}
\label{table:results}
\end{table*}

\begin{figure*}
    \centering
    \includegraphics[width=\textwidth,scale=0.7]{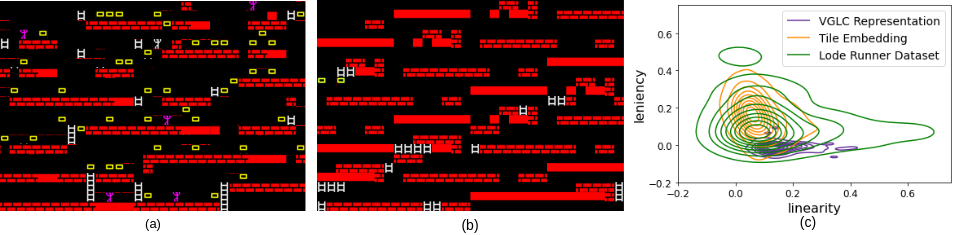}
    \caption {(a) Level generated for Lode Runner by training LSTM on tile embeddings. (b) Level generated for Lode Runner by training LSTM on VGLC tile character representation. (c) Kernel Density Estimation with Linearity and Leniency}
    \label{fig:output_lr}
\end{figure*}

\begin{figure*}[!tbh]
    \centering
     \includegraphics[width=\textwidth]{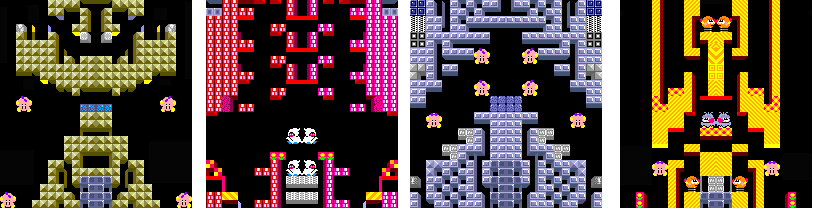} 
     \caption{Levels Generated for Bubble Bobble}
     \label{fig:output_bb}
\end{figure*}

\subsection{Cross-fold Affordances Analysis}
Table \ref{table:results} presents the results of all the evaluation metrics for predicted affordances of unseen game tiles. The most frequent label combination in our dataset is [\textit{'empty','passable'}] accounting for approximately 32\% of the dataset. 
The \textit{Most Frequent Label (MFL) Baseline} indicates the value of our metrics if only the most-frequent label combination is predicted.
We include it as a comparison point in the table and in our discussion of the results below.
For all the metrics, the closer the value is to 1, the better. \textbf{Bold} indicates the highest value and \textit{italic} indicates the lowest value across the test games for our model. 

Exact match ratio (EMR) indicates the percentage of label combinations identified exactly by the model. On average, EMR is 0.29 with a standard deviation of 0.14.
The performance is mainly because the metric is aggressive and does not attribute any value to partially correct predictions. The MFL Baseline achieves 0.32 due to the fact that the label makes up 32\% of the dataset. However, we still outperform it for three of the five games.

We observe stronger performance on example-based measures that evaluate partial correctness. 
The average values observed across all example-based evaluations are better in comparison to our MFL baseline.
However, if we evaluate individual labels, we find lower values. These lower values on label-based metrics is likely due to the poor performance in predicting rare labels, and due to the over-abundance of the most common labels.

In all five games of our dataset, \textit{Solid, Passable, Empty} tiles occupy a majority of the level as compared to other tiles. Concretely, these labels together account for 70.8\% of our training instances. Comparatively higher values on the example-based evaluations than label-based evaluations demonstrate that the model is capable of predicting frequent labels and struggles to predict rare labels such as \textit{Climbable, Collectibles, Element, Block, Wall, Hazard}. For instance, the level design for Legend of Zelda has dungeons composed of \textit{Solid} tiles which our model is good at predicting. 
Hence metrics for Legend of Zelda have higher values than other games. In comparison, Lode Runner has the lowest values for most of the metrics as its levels have a well-proportioned set of tiles including \textit{Enemy, Collectable, Breakable, Solid, Empty, Passable}. It also had the largest set of overall data, and our model clearly struggled when we withheld these training samples. However, our tile embeddings are able to effectively represent Lode Runner levels when trained on this data, as we demonstrate in the next evaluation. 

The table also highlights the effect of different values of $\beta$ and $\gamma$ on the $\alpha$-evaluation scores. Lowering the weight of misclassification errors ($\gamma$) and increasing the weight of missing errors ($\beta$), increases the $\alpha$-evaluation score. This indicates the presence of more misclassification errors and fewer missing labels i.e more False Positives. 

Overall, we find these results to be heartening, as our model outperformed our MFL baseline for seven of our ten metrics, and always performed better than it for at least two games.
This suggests we can approximate affordances on unseen games. 
Additionally, for certain use cases like level generation, getting the exact correct affordances is not required as long as the latent space representations of similar entities are close together.
This is due to the fact that identifying the entities with similar behaviour will ensure they are appropriately handled in terms of placement during level generation. For instance, as long as enemy tiles are grouped together and separate from solid tiles, a secondary model can be trained to place them in appropriate positions. 

\subsection{LSTM for Level Generation-Annotated Game}
Figure \ref{fig:output_lr} gives the results for our second evaluation. We generated 150 levels with each LSTM: one trained on tile embeddings and the other trained with the VGLC representation. Figure \ref{fig:output_lr} shows the Kernel Density Estimation with Leniency and Linearity.
While there is a small section of the plot that the VGLC levels cover that the tile embedding levels do not, overall the levels generated with the tile embeddings representation cover more of the original distribution.
In particular, since the VGLC representation did better in terms of linearity, we expect that the VGLC's hand-authored representation was better able to encode structural knowledge. 
However, the LSTM struggled to model less common elements with it, including enemies and rewards, which can be seen in the Kernel Density Estimation and example level.

\subsection{LSTM for Level Generation-Unannotated Game}
Figure \ref{fig:output_bb} shows the Bubble Bobble levels generated by the LSTM trained on tile embeddings. 
While we note some oddities (floating enemies) the levels overall are of surprisingly high quality, indicating the appropriateness of this approach for generating levels on unseen games. 
We note that these levels were output as a tile embedding and then visualized with Annoy as described above. 
One benefit of our approach is that we naturally model tiles with the same affordances (e.g. solid tiles) with all of the visual variety from the original content, leading to the yellow, blue, and pink structures in the output levels.
Approaches like the VGLC representation cannot due this, and require a secondary process to map tiles with the same affordances to multiple, distinct output tiles. 
To play these levels one would need to map them to in-game objects (which is also necessary for the VGLC representation) or employ the embeddings in a playable, deep neural network-based game \cite{kim2020learning}.

\section{Limitations and Future Work}

Our first evaluation provided evidence that a trained autoencoder can generalize and approximate missing labels on a majority of instances for an unseen game. 
However, there is room for improvement. 
We suggest the following avenues.
First, expanding the affordances, a set of 13 labels is fairly limiting for the model to be able to express any 2D tile-based games. 
If we expand the affordances, we can include additional games of different styles and genres, which will enrich the training corpus.
Our second option would be data augmentation, which is related to the challenge of dataset imbalance. 
While implementing a weighted binary cross-entropy loss provided a significant improvement, the distribution for labels with fewer instances needs to be better generalized. We believe one way to tackle this is by employing sampling techniques.
Third we suggest including the
semantic context of neighboring tiles.
In our current work, we input affordances of the candidate tile and its surrounding visual context. We believe providing the affordances of the neighbouring tiles will help the model to represent a tile in terms of the surrounding visuals and their mechanical relationships. 

In NLP, a language model captures semantic and syntactic relationships between words. Word analogies are often demonstrated using algebraic operations on word vectors. For instance, the example mentioned in the related work, $\vec{king}\ -\ \vec{man}\ +\ \vec{woman}\ \approx\ \vec{queen}$. We identify this as a potential aspect of future research for tile embeddings. Such experiments would validate the capability of the model in capturing the semantic relationships between tiles. We could also leverage this representation for the generation of novel tiles. 
By feeding in affordances and visual noise, the VAE could hopefully transform the noise into a reasonable visual structure that matched the desired affordances.
We hope to explore these directions for future work. 

\section{Conclusion}
In this paper, we presented the first instance of tile embeddings as a common vector representation for 2D tile-based games. We trained an autoencoder to extract embeddings that constitute the visual and semantic information of the tile. Further, we evaluated and presented evidence for approximating affordances for unseen games. We demonstrated that our embeddings can be successfully applied to PCGML tasks like level generation.  

\section*{Acknowledgements}

This work was funded by the Canada CIFAR AI Chairs Program. We acknowledge the support of the Alberta Machine Intelligence Institute (Amii). We acknowledge the support of the Natural Sciences and Engineering Research Council of Canada (NSERC).

\bibliography{Main}

\end{document}